\documentclass{article} 
\usepackage{iclr2016_workshop,times}
\usepackage{hyperref}

\usepackage{times}
\usepackage{helvet}
\usepackage{courier}
\usepackage{graphicx}
\usepackage{caption}
\usepackage{subcaption}
\usepackage{placeins}
\usepackage{multirow}
\usepackage{amssymb}
\usepackage{amsmath}
\usepackage{subcaption}

\DeclareMathOperator*{\argmax}{argmax}

\newcommand*{\argmaxl}{\argmax\limits}
\usepackage{wrapfig}

\title{Convolutional Clustering\\ for Unsupervised Learning}   

\author{
Aysegul Dundar, Jonghoon Jin,  and Eugenio Culurciello \\
Purdue University, West Lafayette, IN 47907, USA \\
\texttt{\{adundar,jhjin,euge\}@purdue.edu} \\
}

\begin{document}
\maketitle

%

\begin{abstract}
The task of labeling data for training deep neural networks is daunting and tedious, requiring millions of labels to achieve the current state-of-the-art results. Such reliance on large amounts of labeled data can be relaxed by exploiting hierarchical features via unsupervised learning techniques.
In this work, we propose to train a deep convolutional network based on an enhanced version of the k-means clustering algorithm, which reduces the number of correlated parameters in the form of similar filters, and thus increases test categorization accuracy.
We call our algorithm \emph{convolutional k-means clustering}. 
We further show that learning the connection between the layers of a deep convolutional neural network improves its ability to be trained on a smaller amount of labeled data.
Our experiments show that the proposed algorithm outperforms other techniques that learn filters unsupervised. Specifically,  we obtained a test accuracy of 74.1\% on STL-10 and a test error of 0.5\% on MNIST.
\end{abstract}
\section{Introduction}

Deep neural networks require massive amounts of data to be trained. In large-scale datasets, supervised methods have been successfully trained over the past few years due to the advances in parallel computing \citep{simonyan2014very, szegedy2014going}. Popular datasets such as ImageNet \citep{deng2009imagenet} contain more than a million labeled samples, and even larger datasets are already sought after by researchers in the field. Further pushing the boundaries, video datasets are becoming increasingly important in the context of deep neural networks for event recognition tasks. In all such cases, labeling is necessary so that a supervised training algorithm can be used. However, the task of labeling data is quite expensive and time-consuming, requiring tedious work. For example, several hundreds of hours were spent to create ImageNet, and thousand of hours may be needed to annotate even the most simple video dataset \citep{ILSVRC15}. To circumvent this problem, the research community recognizes that a large breakthrough lies in the use of unlabeled data, which is freely available in abundant quantities. 

Over the last few decades, extensive research has been dedicated to learning feature hierarchies for deep learning in the context of image understanding. Examples include unsupervised, supervised, and semi-supervised learning. Such deep learning techniques use hierarchy of layers, which use ``filters" to extract multiple input features and ``connections" to combine extracted features together into inputs for the next layer.
In earlier studies in the field, unsupervised pre-training was required for training deep networks by supervised learning methods. Recent advances in Convolutional Neural Networks (\textit{ConvNets}) combined with abundant amounts of labeled data have shown great promises in object recognition tasks to remedy this issue \citep{krizhevsky2012imagenet}.

On the other hand, unsupervised learning algorithms, such as k-means clustering, also increased the number of parameters in the network and achieved state-of-the-art results when labeled data are limited.
Although unsupervised learning techniques using k-means algorithm were commonly used to train filters in several studies \citep{coates2011selecting,bo2013unsupervised}, the network encoding structures present many similarities with ConvNets, such as the use of convolution and pooling in each layer.

The main differences between ConvNets and unsupervised learning techniques based on k-means applied to image recognition are the number of layers (depth) and the number of filters (width) at each layer, and the connections among layers. 
ConvNets improve accuracy by increasing network depth and width. Recent studies show that, significant performance of ConvNets was a result of the increased depth \citep{zeiler2014visualizing}.
By contrast, unsupervised learning algorithms for deep networks were not able to scale to the same depth as conventional ConvNets.
Therefore, recent unsupervised studies use large network width and two-to-three layers with diminishing returns \citep{coates2011selecting}.
In this work, we demonstrate that learning the connections between the layers of deep neural networks plays a crucial role in improving the performance of unsupervised techniques.

While early work of ConvNets used to rely on a `non-complete' connection scheme  \citep{lecun1998gradient} to keep the number of connections within reasonable bounds, the trend has changed to fully-connected layers in order to exploit the benefits of parallel computing \citep{krizhevsky2012imagenet, simonyan2014very}.
Fully-connected layers perform a lot of potentially unnecessary operations because they connect every feature of the previous layer to every feature of the next.

In this study, a refined version of an unsupervised clustering algorithm that allows the filters to learn diverse features has been proposed. This is achieved by preventing the algorithm from learning redundant filters that are basically shifted version of each others as explained in detail in Section  \ref{sec:kmeans}. 
Another major contribution of this work is that we learn sparse connection matrices between layers by forcing sparser group of features to map into the feature of next layer which has been explained in Section \ref{sec:connections}.
We show that the convolutional k-means clustering algorithm can provide comparable mid-level feature hierarchies to the supervised networks with improved connection learning.

\section{Related Work}

In recent years, there has been an increasing interest to learn ConvNets filters using unsupervised learning either in pre-training or when specifying the filter values.
Earlier work suggested to use sparse coding and sparse modeling at patch level ignoring the fact that filters would be used in a convolutional manner \citep{poultney2006efficient,zeiler2010deconvolutional}.
Such approaches result in duplicated filters that are simply shifted versions of each others.
To address this problem, convolutional Restricted Boltzmann Machines trained with contrastive divergence \citep{lee2009convolutional} and convolutional sparse coding \citep{kavukcuoglu2010learning} methods were proposed.

Filters using k-means algorithm have gained significant attention in recent studies because of its simplicity and its competitive results when combined with the right pre-processing and encoding scheme \citep{coates2011importance,lin2014stable}.
In these studies, filters trained with the k-means algorithm are applied in a convolutional manner over the input maps to extract useful features. However, there has not been any attempt to reduce the redundancy between the filters learned with this algorithm, a problem that hampers efficiency and accuracy.

As in almost all statistical learning problems, curse of dimensionality is a known issue in deep neural networks.
In particular, studies show that  k-means performs poorly after the first layer \citep{coates2011selecting}. 
The number of filters of the first layer have low dimensions, on the other hand, the subsequent layers increase the number of network parameters exponentially.
As an example to the curse of dimensionality problem, if we have $32\times32$ RGB images, and we train $96$ $3\times5\times5$ pixel filters in the first layer and convolve them with input images, we will get $96\times28\times28$ feature maps as output.
If we want to train fully connected filters in the second layer (as in the first layer filters), we would need to train $96\times5\times5$ filters.
The k-means algorithm fails to extract distinctive features and works poorly in such a high dimension. 
Therefore, for the mid level features, a smaller receptive field than fully connected layer should be preferred \citep{coates2011selecting}.
In the early work of ConvNets, \citet{lecun1998gradient} used parsimonious (not fully-connected) connection schemes to keep the number of connections within reasonable bounds and to force a break of symmetry in the network. Since different feature maps are fed with different input sets, the system is forced to extract different features. 
In techniques that use unsupervised algorithms, random connection \citep{culurciello2013analysis}, and grouping similar features \citep{coates2011selecting} have been proposed; these results added additional layers and provided some improvement but not as significant as the ones obtained with supervised deep network.


In this work, we address the aforementioned problems
by devising an optimized learning algorithm that avoids replication of similar filters. Since the filters will be used in convolutional operation, shifted versions of filters do not provide additional information to the feature hierarchy, and therefore should be avoided. 
We further propose to learn the connections between layers via supervised learning in the context of ConvNets. The connection setup uses 1D convolution across channels which is equivalent to the operation denoted as \emph{mlpconv} layers in \citet{lin_2013_nin}.
This layer has been used to enhance the abstraction ability of the local model in \citet{lin_2013_nin}, and to decrease the dimension of modules as well as to remove the computational bottlenecks in \citet{szegedy2014going}.

\section{Learning Filters}
\label{sec:kmeans}
\begin{figure*}
  \centering
  \begin{subfigure}[b]{0.5\textwidth}
    \centering
    \includegraphics[width=0.9\textwidth]{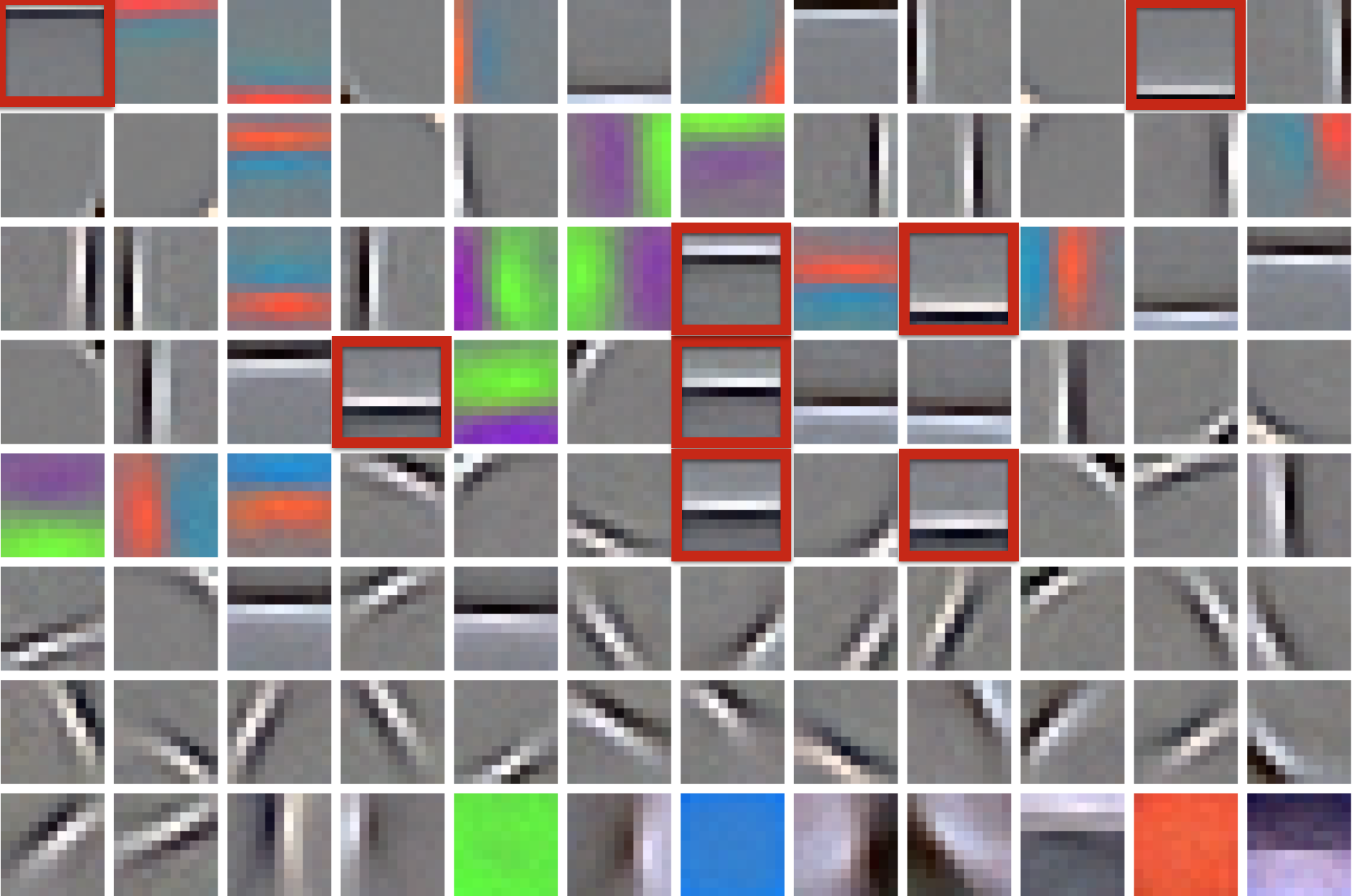}
    \caption{ k-means}
    \label{fig:filters-kmeans}
  \end{subfigure}%
  \begin{subfigure}[b]{0.5\textwidth}
      \centering
    \includegraphics[width=0.9\textwidth]{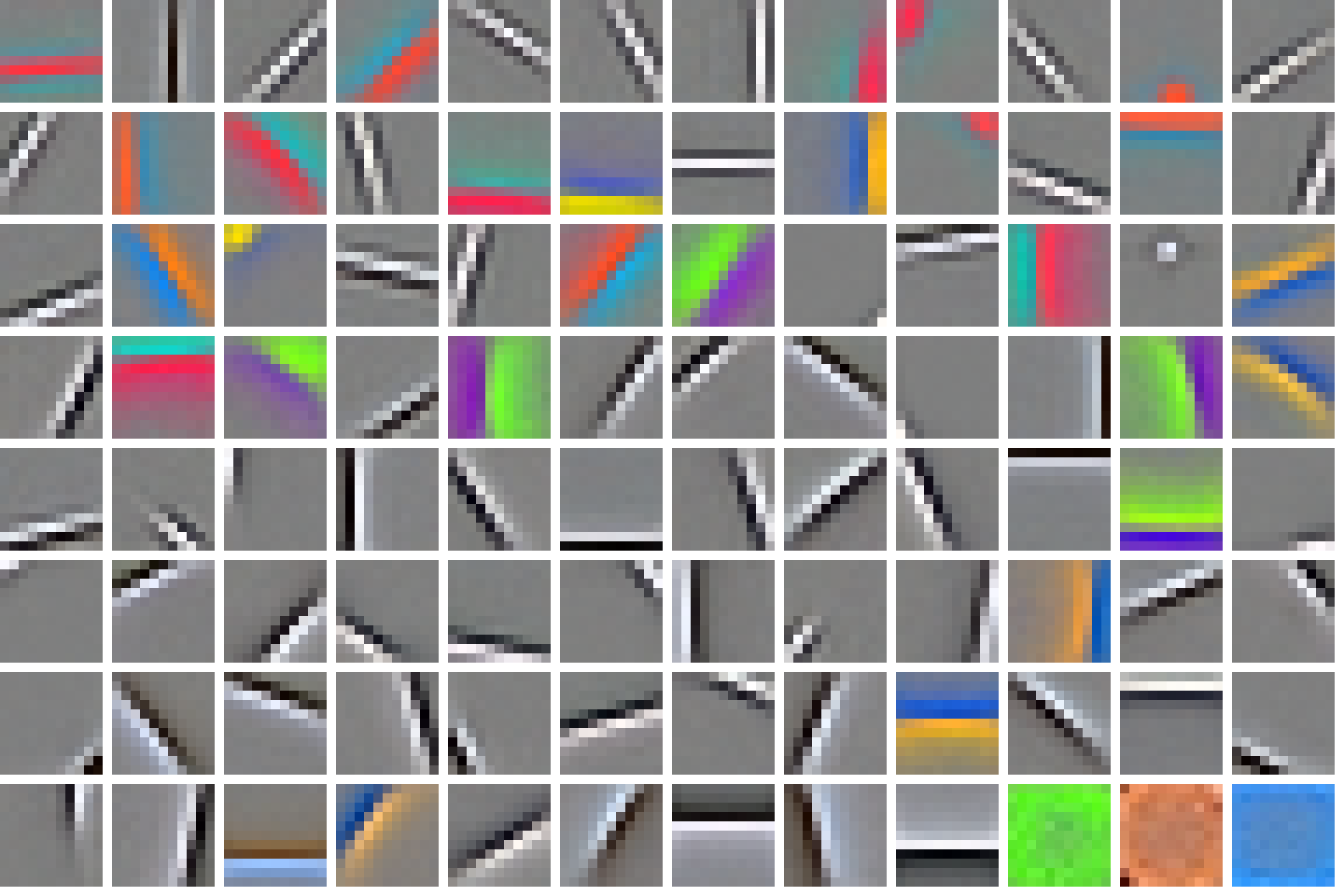}
    \caption{convolutional k-means}
    \label{fig:filters-conv}
  \end{subfigure}
  \caption{Filters trained on the STL-10 dataset with k-means and convolutional k-means.
  Filters are sorted by variance in descending order. 
  While convolutional k-means learns unique features, the k-means algorithm introduces redundancy in filters.
  The duplicated features for horizontal edges are highlighted in red.}
  \label{fig:filters}
\end{figure*}

\subsection{Learning Filters with k-means}

Our method for learning filters is based on the k-means algorithm.
The classic k-means algorithm finds cluster centroids that minimize the distance between points in the Euclidean space. In this context, the points are randomly extracted image patches and the centroids are the filters that will be used to encode images. From this perspective, k-means algorithm learns a dictionary $D \in \mathbb{R} ^{n \times k}$ from the data vector $w^{(i)} \in \mathbb{R} ^{n}$ for $i=1,2,...,m$.
The algorithm finds the dictionary as follows:

\begin{equation}
\begin{aligned}
 & s_j^{(i)} \ \ \ :=
  \begin{cases}
   D^{(j)^T}w^{(i)} & \text{if } j = \argmaxl_{l}  \left|D^{(l)^T}w^{(i)}\right|, \\
   0      & \text{otherwise,}
  \end{cases}\\
  & {D \ \ \ \ \ := WS^T+D},\\
  & {D^{(j)} \ := \frac{D^{(j)}}{||D^{(j)}||_2}}\, , 
 \end{aligned}
  \label{eq:kmeans}
\end{equation}
where $s^{(i)} \in \mathbb{R}^k$ is the code vector associated with the input $w^{(i)}$, and $D^{(j)}$ is the $j$'th column of the dictionary $D$. The matrices $W \in  \mathbb{R}^{n \times m}$ and $S\in \mathbb{R}^{k \times m}$ have the columns $w^{(i)}$ and $s^{(i)}$, respectively.
$w^{(i)}$'s are randomly extracted patches from input images that have the same dimension as the dictionary vectors, $D^{(j)}$. 

Described learning scheme trains the centroid of each cluster at the patch level, however, in ConvNets, filters are applied to images in a convolutional manner.
As observed in Figure \ref{fig:filters-kmeans}, many of the centroids from the k-means training have almost the same orientation and they are shifted  versions of each other in space. Therefore, after the convolution operation, they will produce redundant feature maps at neighboring locations.
In the next section, we explain the proposed modifications of the k-means algorithm (\textit{convolutional k-means}) that alleviates this problem.

\subsection{Learning Convolutional Filters with k-means}

In order to reduce the redundancy between filters at neighboring locations, we propose a new input patch extraction method. This method significantly reduces the redundancy in centroids produced by the k-means algorithm and keeps only the essential basis for them.
The standard k-means algorithm extracts random patches from input images whose dimensions match those of the centroids.
By contrast, the proposed method uses larger windows as inputs to decide which patch to extract for clustering.

The windows are chosen to be two times bigger than the filter size and randomly selected from the input images. The centroids of the k-means algorithm convolve the entire window to compute a similarity metric at each location of the extracted area.
The patch which corresponds to the biggest activation from the window is meant to be the most similar feature to the centroid (given that ConvNets have translation invariance).
Finally, the patch at that specific location (biggest activation) is extracted from the window and it is assigned to the corresponding centroid. The modified dictionary learning can be written as follows:

\begin{equation}
\begin{aligned}
 & s_j^{(i)} \ \ \ :=
  \begin{cases}
   D^{(j)^T}w^{(i)}_{(x, y)} & \text{if } (j, x, y) = \argmaxl_{(l, m, n)} \left|D^{(l)^T}w^{(i)}_{(m, n)}\right|, \\
   0      & \text{otherwise,}
  \end{cases}\\
  & {D\ \ \ \ \ := W_{(x,y)}S^T+D},\\
  & {D^{(j)} \ := \frac{D^{(j)}}{||D^{(j)}||_2}}\, ,
 \end{aligned}
  \label{eq:kmeans-conv}
\end{equation}

where $D^{(j)}$ is the $j$'th column of the dictionary that corresponds to a $c\times s\times s$ 3D filter kernel  and $w^{(i)}$ is the window with size $c\times 2s\times 2s$.
$x$ and $y$ are the top-left location index of the input patch, and $w^{(i)}_{(x, y)}$ is the extracted patch from the location $(x, y)$ with size $c\times s\times s$.

When these correlated filters are removed, there is more room for new filters to learn additional features.
The filters that are trained with both k-means and convolutional k-means algorithms are presented in Figure \ref{fig:filters}.
As can be observed from Figure \ref{fig:filters-kmeans}, filters that are trained at the patch level with k-means algorithm have similar features but at different locations within a patch.
As an example and also highlighted in red, there are many horizontal filters that are replicas of each other at different heights.
By contrast, the filters that are trained with the convolutional k-means algorithm are significantly more diverse, as can be seen in Figure \ref{fig:filters-conv}.

\begin{figure*} 
  \centering
  \begin{subfigure}[b]{0.5\textwidth}
   \includegraphics[width=\textwidth]{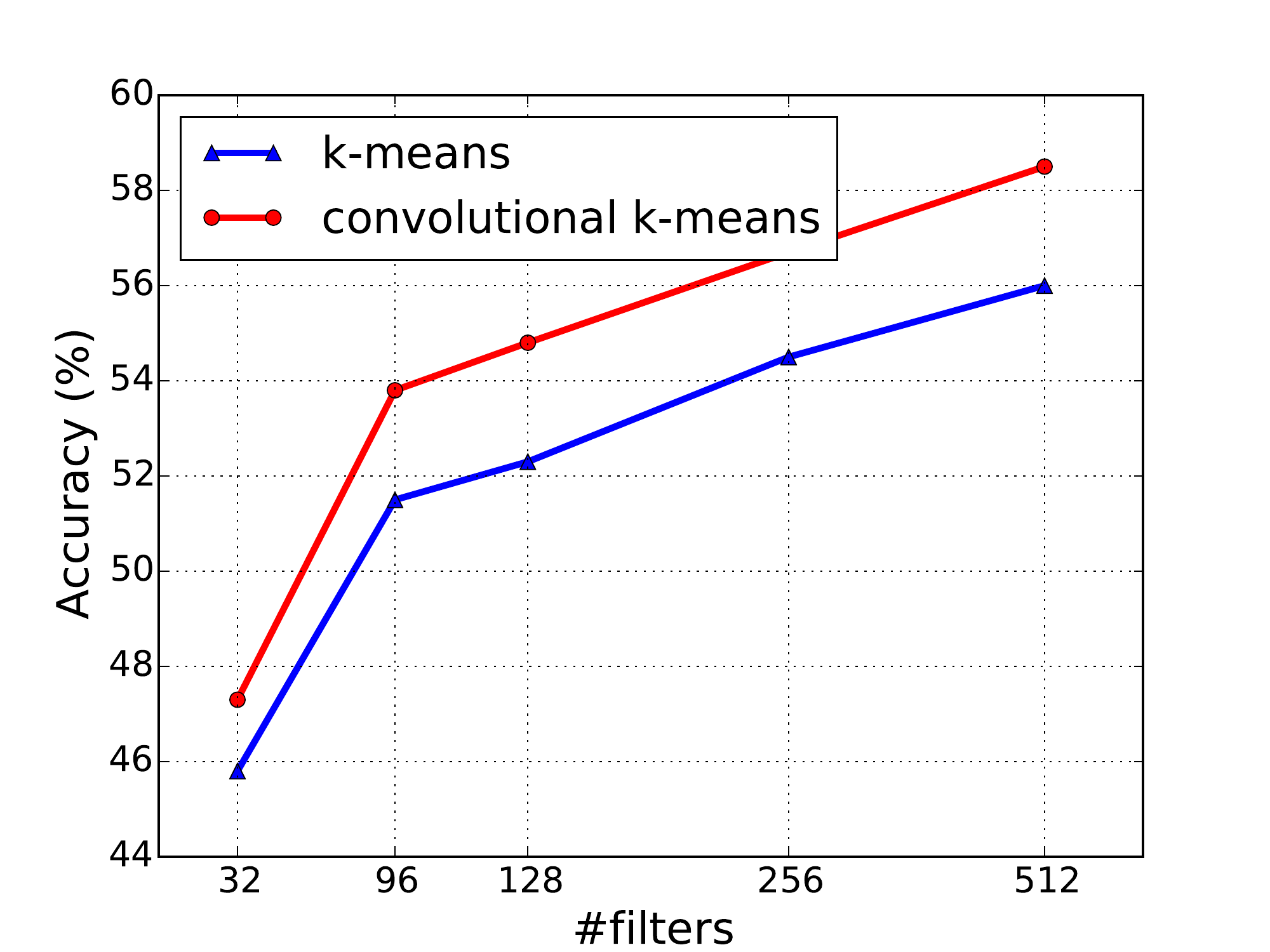}
    \caption{Accuracy versus number of filters.}
  \label{fig:acc-features}
   \end{subfigure}%
  \begin{subfigure}[b]{0.5\textwidth}
     \includegraphics[width=\textwidth]{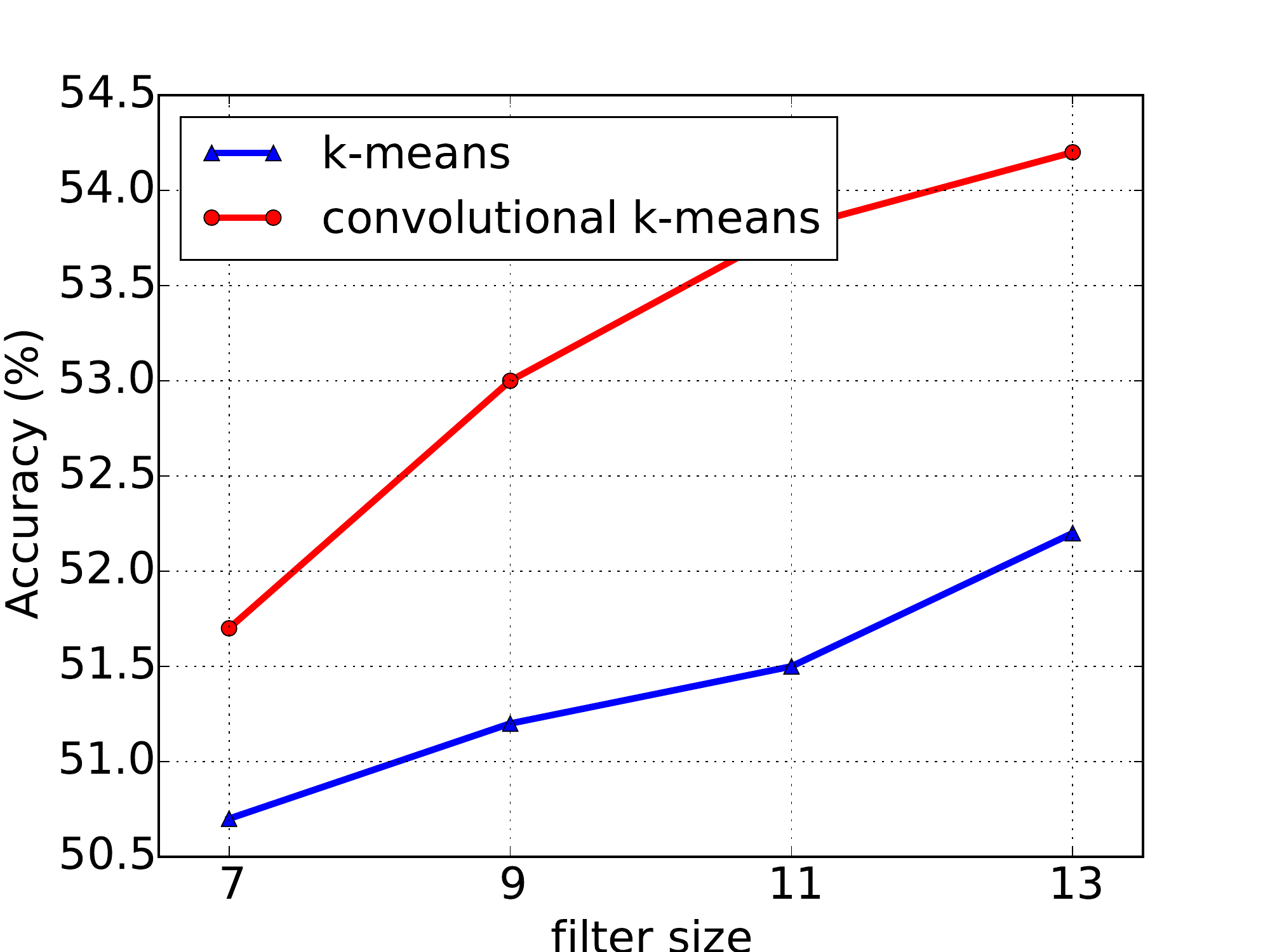}
    \caption{Accuracy versus filter size.}
  \label{fig:acc-size}
   \end{subfigure}
  \caption{
    Comparisons of accuracy on the STL-10 dataset with filters that are trained by k-means and convolutional k-means.
    These tests use a single layer network and the sizes of filters are fixed to $11\times11$ for (a) while the number of filters is set to $96$ for (b).
  }
  \label{fig:accuracies}
\end{figure*}

\subsection{Experimental Results of Single Layer Network}\label{sec:oneLayer}

We run experiments of a single layer network to analyze the effect of convolutional k-means.
In our experiments, we use the STL-10 dataset that contains $96\times96$ RGB images in $10$ categories \citep{coates2011analysis}.
This dataset has $500$ images per class for training and $800$ for testing.
Additionally, it includes $100,000$ unlabeled images for unsupervised learning algorithms which are extracted from similar but broader distribution of images.
For learning the filters with k-means clustering, only unlabeled data are used.
For the training of filters with k-means and convolutional k-means, patches are randomly extracted from the raw images.
Then, standard pre-processing, such as global contrast normalization and ZCA-whitening, is applied to the extracted patches.

For the encoding scheme, we only apply global contrast normalization to the input images. 
We fix the first layer filters of ConvNet with the trained dictionary by k-means and convolutional k-means.
We reduce the dynamic range of the trained filters by dividing the filter values by a constant that is determined by cross-validation.
In our experiments, we use the STL-10 dataset without any downsampling. However, the convolutional layer is applied with a stride of $4$, which effectively reduces the dimension in the first layer.
This layer is followed by a max-pooling operation, which reduces the dimension to $\text{K}\times2\times2$, where K is the number of filters that are used in convolutional layer.
After pooling, rectification linear unit (ReLU) activation function is used;  similar to recent works in ConvNets \citep{krizhevsky2012imagenet}.
Note that to compare the effectiveness of filters that are trained with these two algorithms, 
we set a large pooling size which would decrease the dimension to $\text{K}\times2\times2$, and train a single layer classifier. 
In the experiments, we use a learning rate of $0.1$ with a momentum rate of $0.9$.

In Figure \ref{fig:accuracies}, we compare the k-means algorithm with our convolutional k-means approach. In Figure \ref{fig:acc-features}, we fix the filter size to $3\times11\times11$ and change the number of filters. We observe that the increase in the number of filters provides us with higher performance for both algorithms, however, filters that are learned with convolutional k-means always outperform the ones with k-means algorithm. Note that to achieve a similar level of accuracy, such as $54\%$, the number of required filters for our approach is smaller than half of those for k-means.  
In Figure \ref{fig:acc-size}, we fix the number of filters to $96$ and vary the size of the filters. Our approach outperforms k-means for all filter sizes.

\section{Learning Connections}
\label{sec:connections}

We also study a way to learn connections from one network layer to the next. Such connections are of extreme importance as creating groups of feature maps from which the following layer learns new features.
While fully-connected layers make use of all the features of the previous layer into the next one, we use non-complete connection  \citep{lecun1998gradient}, which are more efficient in computation. These non-complete connections use multiple groups, each including a limited portion of the previous layer features. We use a sparse connection matrix that limits the local receptive field. Consequently, we can avoid the poor performance of the k-means algorithm when the input data are high dimensional \citep{coates2011selecting}.

\begin{figure*}
  \centering
  \includegraphics[width=1\textwidth]{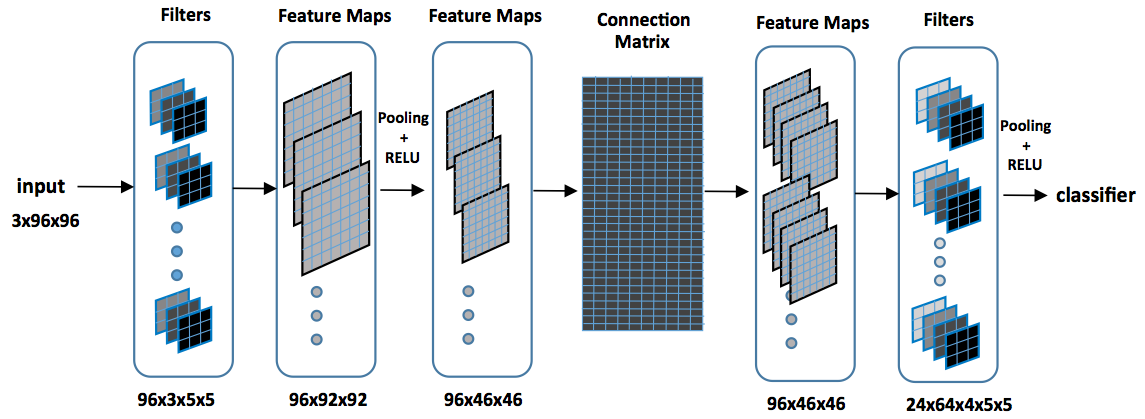}
  \caption{
    Learning Connections setup. 
    The setup network includes a connection matrix and a convolutional layer with a predefined non-complete connection scheme.  
     First, the network with randomly initialized connection matrix is trained with supervised learning to learn the correct connection weights. Second, using the trained matrix, the next-layer filters are learned with convolutional k-means, as performed in the previous layer.
  }
  \label{fig:struct}
\end{figure*}

Our method makes use of supervision with limited data while learning the connection weights between layers.
The connections are described by a fully connected weight matrix that pools over the feature maps.
Therefore, a single value in the weight matrix reflects how important that feature is for the corresponding group.
To learn the relation between maps and organize them as groups (i.e., to define their weights), we add a convolutional layer with a predefined non-complete connection as illustrated in Figure \ref{fig:struct}.
We attach a linear classifier after the convolutional layer and train the system using a backpropagation algorithm.
The intuition of this setup is that since new filters are learned from groups of features and each weight matrix pools over features whose output is in a pre-determined group, the weight matrix is actually forced to learn the \textit{proper} connections between the input feature maps and pre-determined groups by training.
Therefore, even though we pre-defined the connections of the convolutional layer, the weight matrix provides the network with flexibility to define the connections in practice. 

Note that the weight matrix that pools over feature maps can be considered as a 1D fully connected convolutional layer with enabled bias.
This new approach allows us to limit the local receptive area for the k-means algorithm. It further allows us to create new complex and learnable interactions of cross channel information through the trained weights.
After we learn the connections via supervised learning, 
we remove the learned filters from the network and only keep the connection matrix. This is because our goal is to learn the filters with unsupervised technique, using minimal labeled data, and to avoid overfitting.
After these steps, our convolutional k-means algorithm is applied again on the pre-trained connection matrix to learn the filters for the next layer, and the algorithm no longer suffers from the curse of dimensionality.
Details and experimental results are presented in the next section. 

\begin{table}[hbt!]
\caption{Classification accuracy on STL-10 testing set with 2 layer networks.}
\label{stl-experiment1}
\begin{center}
\begin{tabular}{cccc}
\multicolumn{1}{c}{\bf First Layer} &\multicolumn{1}{c}{\bf Connection} &\multicolumn{1}{c}{\bf Second Layer} &\multicolumn{1}{c}{\bf Accuracy}
\\ \hline \\
Supervised                  &Supervised               &Supervised             &$62.5\%$ \\
Unsupervised              &Random                   &Supervised             &$64.7\%$ \\
Unsupervised              &Random                   &Unsupervised         &$65.4\%$ \\
Unsupervised              &Supervised               &Supervised            &$66.2\%$ \\
Unsupervised              &Supervised               &Unsupervised        &$67.1\%$ \\

\end{tabular}
\end{center}
\end{table}

\subsection{Experimental Results of Multi-layer Networks}\label{sec:multilayers}

We conduct experiments combining (a) supervised and unsupervised learned filters and (b) supervised learned and random connections between layers. These experiments are designed to analyze the importance of learning connections. We set up a 2 layer network. The first layer has $96$ filters of size $13\times13$.
The convolutional layer is applied with a stride of $4$ and followed by ReLU.
Between the first layer and second layer feature extractors, we pre-define groups as 4 consecutive feature maps, which results in  $96/4=24$ groups.
From each group, we learn 64 filters. The size of second layer filter is chosen to be $4\times5\times5$, 4 comes from the choice of pre-defined non-complete connection scheme.
After the convolution with the filters, we apply a pooling operation of $6\times6$ to  decrease the dimensions. 
ReLU activation function follows the max-pooling operation.
We use a linear classifier with $2$ layers with a hidden neuron of $512$ and interleaved with dropout \citep{hinton2012improving}.

For the \textit{unsupervised learning} of filters, 
we apply convolutional k-means algorithm to the unlabeled data.
\textit{Random connection} refers to the case where the first layer filters are connected to the second layer with a pre-defined connection scheme.
\textit{Supervised connection} refers to the case where  we train a supervised connection matrix of $96\times96$ before the pre-defined connection scheme. 
The second layer filters and the connection matrix are trained together, this forces the connection matrix to organize the feature maps such that the each group contains features that should be combined together.
To learn the second layer filters in unsupervised manner with supervised connections, we fix the supervised trained connection matrix and train local filters for each group with convolutional k-means.

\begin{figure*} 
  \centering
  \begin{subfigure}[b]{0.5\textwidth}
    \centering
    \includegraphics[width=\textwidth]{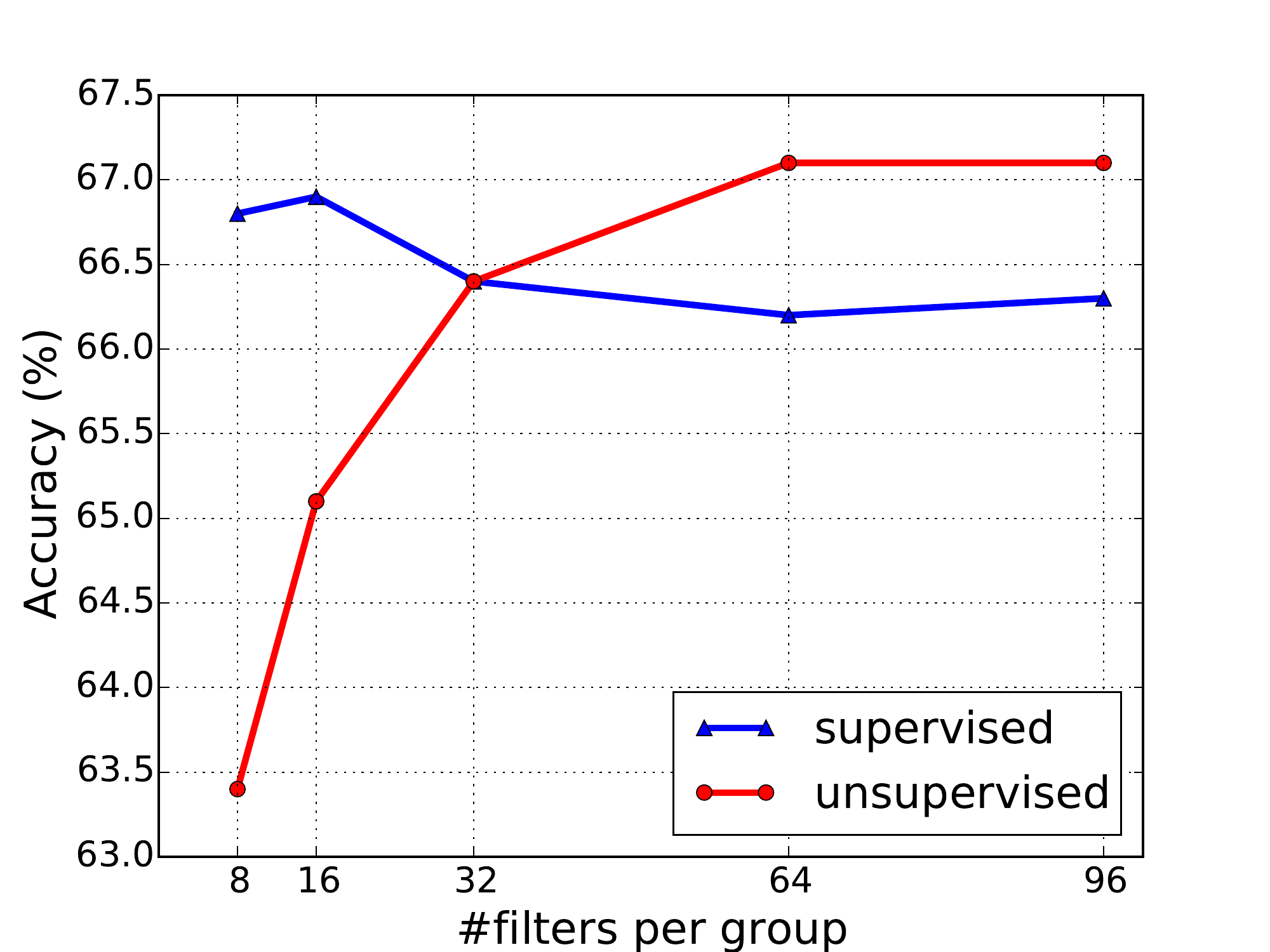}
    \caption{ 2 layer network}
    \label{fig:2layer}
  \end{subfigure}%
  \begin{subfigure}[b]{0.5\textwidth}
      \centering
    \includegraphics[width=\textwidth]{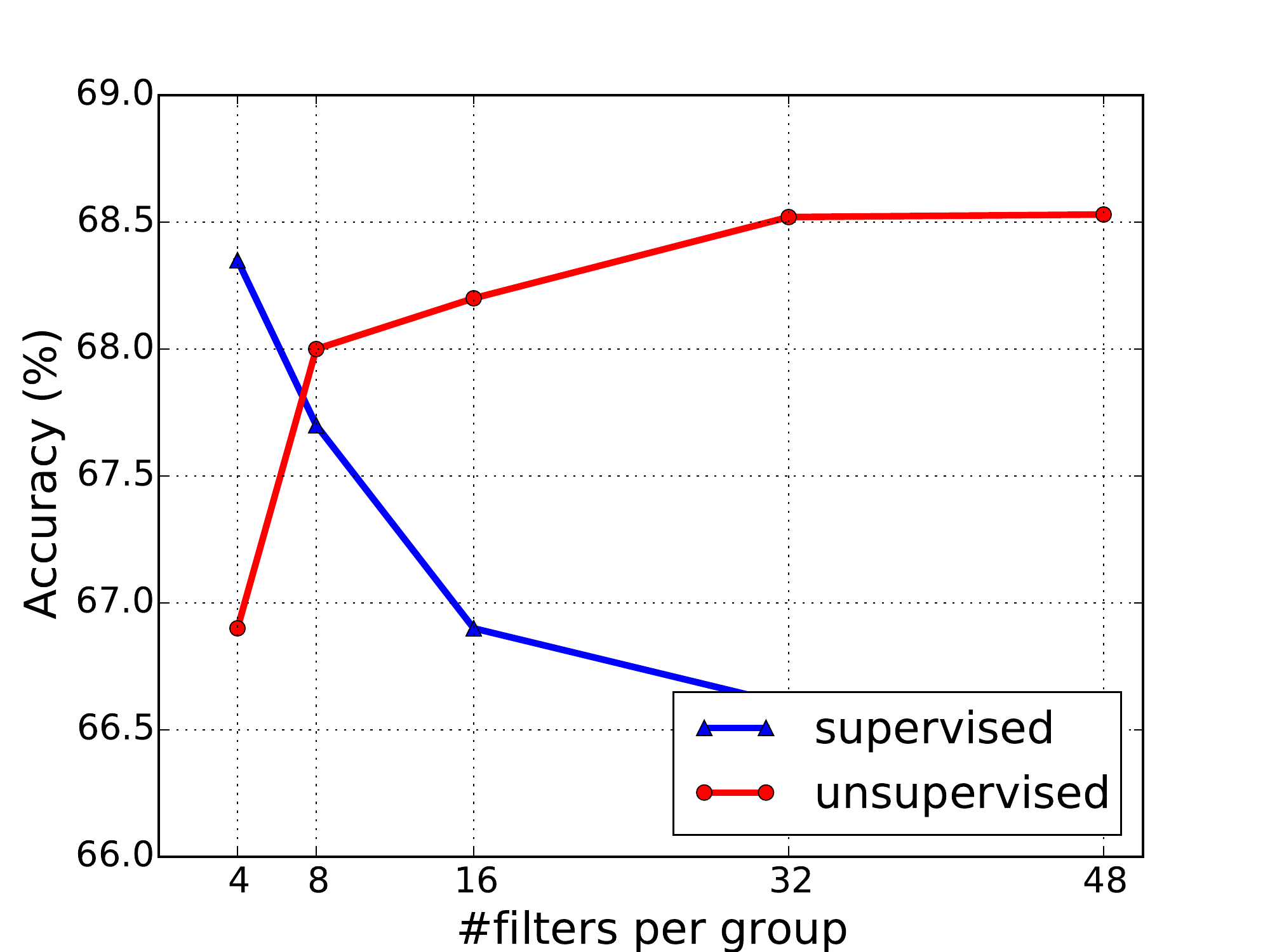}
    \caption{3 layer network}
    \label{fig:3layer}
  \end{subfigure}
  \caption{
    Performance comparisons of two and three layer networks with different learning methods on the STL-10 dataset.
    Supervised denotes that the corresponding layer trained via standard backpropogation, and unsupervised (this work) convolutional k-means filters and learned connections.
  }
  \label{fig:networks}
\end{figure*}

Table \ref{stl-experiment1} shows the results of these experiments. First of all, training the whole network with supervised learning yields lower accuracy, as expected, due to overfitting to the limited number of labeled data.
The unsupervised learning for the first layer provides a large performance increase over a fully supervised network.
In our experiments, learning the connections in a supervised manner boosts the performance for each case although the unsupervised learning still yields better performance. 
Furthermore, in Figure \ref{fig:2layer}, we analyze the effect of the supervised and unsupervised learning of filters in the second layer.
The unsupervised (k-means algorithm) and supervised (backpropogation algorithm) learning algorithms show different characteristics as we increase the number of filters in the second layer.
K-means learning algorithm requires inclusion of increasing the number of filters to yield comparable results with the supervised backpropogation algorithm.
Despite the fact that the supervised algorithm can more efficiently represent the data with fewer filters, it loses accuracy and overfits to the training set when the number of filters is increased.
By contrast, the unsupervised algorithm (convolutional k-means) performs poorly with a low number of filters.
This difference can be because the supervised algorithm is learning the discriminative features, whereas k-means learning algorithm learns all kind of common occurred features.

Finally, we extended the depth of the network to three to analyze whether the observed behavior continues with bigger networks.
Using a configuration similar to the second layer, we add a third layer which includes a connection matrix that represents the connections and another convolution layer with non-complete connections.
The connection matrix in this case decreases the dimension (size $1536 \times 678 $) in a similar manner as \citet{szegedy2014going}; this alleviates the computational bottlenecks.
The other cascaded convolution layer groups each four feature maps and learns filters with dimensions $4\times3\times3$. This is followed by a ReLU activation function and max-pooling $4\times4$ to decrease the dimensions.

In Figure \ref{fig:3layer}, we analyze the effect of supervised and unsupervised learning of filters in the third layer.
The results present a similar behavior as in the second layer counterpart.
The k-means algorithm requires to increase the number of filters to yield comparable results than the supervised backpropogation algorithm.
By contrast, the performance of the unsupervised method can be increased further by concatenating the representations computed at different layers as an image feature vector for use in classification.
Instead of just using the last layer output to feed the classifier, we concatenate intermediate layer outputs to feed the classifier in our final results. The improvement is possible because our model does not overfit, as seen in other works \citep{coates2011selecting,lin2014stable}.

\begin{table*}[t!]

\caption{Classification accuracy on STL-10.}
\label{stl-results}
 \begin{subtable}{\linewidth}
 \centering
 \caption{Algorithms that learn the filters unsupervised.}
\begin{tabular}{cccc}
\multicolumn{1}{c}{\bf Algorithm} &\multicolumn{1}{c}{\bf Test Accuracy}
\\ \hline
\cite{coates2011importance} (1 layer)                                                       &$59.0  \%$ \\        
\cite{coates2011selecting}   (3 layers + multi dict.)                                    &$60.1 \% $\\         
 \cite{hui2013direct}   (3 layers )                                                                &$63.7 \%$ \\         
\cite{bo2013unsupervised}  (2 layers + multi dict.)                                    &$64.5 \%$ \\         
\cite{lin2014stable}  (3 layers + multi dict.)                                                &$67.9 \%$  \\

 \hline
\textbf{This work (2 layers + multi dict.)}                                                                 &$71.4 \%$ \\       
\textbf{This work (3 layers + multi dict.)}                                                                 &$74.1\%$ \\       
 \hline

\end{tabular}

\end{subtable}

\vspace{4mm}

 \begin{subtable}{\linewidth}
 \centering
 \caption{Supervised and semi-supervised algorithms.}
\begin{tabular}{cccc}
\multicolumn{1}{c}{\bf Algorithm} &\multicolumn{1}{c}{\bf Test Accuracy}
\\ \hline

 \cite{swersky2013multi}                                                                             &$70.1\%$\\
  \cite{paine2014analysis}(unsupervised pre-training)                                &$70.2\%$\\
 \cite{hoffer2014deep} (triplet network)                                                       &$70.7 \%$ \\
 \cite{dosovitskiy2014discriminative} (exemplar convnets)                         &$72.8\%$ \\
  \cite{zhao2015stacked} (semi-supervised auto-encoder)                         &$74.3\%$ \\
   \hline

\end{tabular}
\end{subtable}
\end{table*}

\section{Final Classification Results}
\label{sec:results}

Finally, we compare our method against published state-of-the-art competing methods on the STL-10 and MNIST datasets.
For this comparison, we mainly focus on algorithms that learn filters in an unsupervised manner. 
Multi-dictionary approach \citep{coates2011selecting,lin2014stable} is the concatenation of the representations that are computed at different layers (i.e., output values) as an image feature vector. 
We use the same learning parameters, pre-processing and encoding scheme as were used in our other experiments (Section \ref{sec:oneLayer}). 

\subsection{STL-10}
For the final classification results, we use networks based on our two and three layer networks experiments. However, we increase the network size by replacing the stride in the first layer with a $2\times2$ max-pooling which increases the accuracy. We further increase the accuracy by the multi-dictionary approach.
In detail, for the two layers with multi-dictionary network, we use a similar network from two layer experiment where we learned 64 filters from each 24 groups in the first layer output.
We concatenate this network with a one layer network with 512 filters. The one layer network also includes ReLU activation and max-pooling to decrease the dimension of the output to $512\times4\times4$.
For the three layers with multi-dictionary network, we use the network from three layer network experiment  where we created 32 filters from 192 groups. We also concatenate this network with a one layer network with 512 filters.
As in previous comparisons, the linear classifier uses 2 layers with a hidden layer of $512$ and interleaved with dropout \citep{hinton2012improving} with a rate of $0.5$. 
As observed in Table \ref{stl-results}, the two layer network with multi-dictionary achieves an accuracy of $71.4 \%$. Note this value is significantly higher than all of the previously unsupervised learning algorithm work, while the network is an order of magnitude smaller (in number of parameters) than the networks used in \citep{coates2011selecting,lin2014stable}. With an additional layer, our algorithm achieves an accuracy of $74.1 \%$.

\subsection{MNIST}
We run a series of experiments on MNIST dataset. For testing, we use the standard $10,000$ test samples and use different sizes of labeled data for supervised trainings as presented in Table \ref{mnist-results}. The training data are randomly sampled from the entire dataset by making sure that each labels are uniformly distributed.
For the unsupervised filter learning algorithm, we use the whole dataset, whereas for training the connections and the classifier, we only use the randomly extracted samples.
We use the same two-layer network that was used on the STL-10 dataset, except this time we decrease the size of the hidden layer in the linear classifier to $256$ and the concatenated one layer network has 96 filters. The experimental results for this dataset can be found in Table \ref{mnist-results}.

\begin{table*}[ht!]

\caption{Classification error on MNIST.}
\label{mnist-results}
 \begin{subtable}{\linewidth}
 \centering
 \caption{
 Algorithms that learn the filters unsupervised.}
\begin{tabular}{ccccc}
\multicolumn{1}{c}{\bf Algorithm}   &\multicolumn{1}{c}{\bf 600} &\multicolumn{1}{c}{\bf 1000} &\multicolumn{1}{c}{\bf 3000} &\multicolumn{1}{c}{\bf All}
\\ \hline 
\cite{zhao2015stacked} (auto-encoder)                                   &$8.4  \%$       &$6.40  \%$         &$4.76 \%$         &- \\
\cite{rifai2011contractive} (constractive auto-encoder)            &$6.3  \%$       &$4.77  \%$         &$3.22  \%$       &$1.14  \%$\\
\textbf{This work (2 layers + multi dict.)}                                  &$2.8 \%$        &$2.5  \%$          &$1.4 \%$           &$0.5 \%$ \\
 \hline

\end{tabular}
\end{subtable}

\vspace{4mm}

 \begin{subtable}{\linewidth}
 \centering
 \caption{Supervised and semi-supervised algorithms.}
\begin{tabular}{ccccc}
\multicolumn{1}{c}{\bf Algorithm} &\multicolumn{1}{c}{\bf 600} &\multicolumn{1}{c}{\bf 1000} &\multicolumn{1}{c}{\bf 3000}  &\multicolumn{1}{c}{\bf All}
\\ \hline
 \cite{lecun1998gradient} (convnet)                                                         &$7.68\%$                       &$6.45\%$                     &$3.35\%$ \\  
  \cite{lee2013pseudo} (psuedo-label)                                                     &$5.03\%$                       &$3.46\%$                     &$2.69\%$                    &- \\  
    \cite{zhao2015stacked} (semi-supervised auto-encoder)                   &$3.31\%$                       &$2.83\%$                     &$2.10\%$                     &$0.71\%$\\
  \cite{kingma2014semi} (generative models)                                         &$2.59\%$                       &$2.40\%$                      &$2.18\%$                    &$0.96\%$\\
  \cite{rasmus2015semi} (semi-supervised ladder)                                 &-                                     &$1.0\%$                       &-                                   &- \\
   \hline

\end{tabular}
\end{subtable}
\end{table*}

\section{Conclusions}

We have presented a novel framework that combines the strengths of an unsupervised clustering algorithm, k-means, and Convolutional Neural Networks when very few labeled data are available.
Our framework modifies the k-means clustering algorithm so that, when used with ConvNets, it learns filters that are less redundant at neighboring locations.
In addition, we proposed a supervised learning setup to learn the \textit{proper} connections between layers.
The idea of local connectivity applied to ConvNets mitigates the curse of dimensionality in filter learning and makes the algorithm scalable.
Moreover, the proposed framework removes the necessity of data whitening on any of the layers including the input during the encoding phase (whitening is applied while learning the dictionary); which makes the encoding stage very simple compared to the others \citep{coates2011selecting,hui2013direct}.
Our experiments show that the proposed algorithm performs better than the state-of-the-art among the techniques that learn deep neural network filters unsupervised.

\subsubsection*{Acknowledgments}
Work supported by Office of Naval Research (ONR) grants 14PR02106-01 P00004 and MURI N000141010278.

\bibliographystyle{iclr2016_workshop}
\bibliography{kmeans-nips2015}

\end{document}